\pdfoutput=1

\documentclass[11pt]{article}

\usepackage[]{naacl2021}

\usepackage{times}
\usepackage{latexsym}

\usepackage[T1]{fontenc}

\usepackage[utf8]{inputenc}

\usepackage{microtype}

\usepackage{amsmath}
\usepackage{amssymb}
\usepackage{amsthm}
\usepackage{bbm}
\usepackage{graphicx}
\usepackage{makecell}
\usepackage{float}
\usepackage{comment}
\usepackage{enumitem}

\DeclareMathOperator{\EE}{\mathbb{E}}
\global\long\def\given{\mid}

\usepackage[nameinlink]{cleveref}
\creflabelformat{equation}{#1#2#3}
\crefname{equation}{eq.}{eqs.}
\Crefname{equation}{Eq.}{Eqs.}
\Crefname{section}{\S}{\S}
\Crefname{lemma}{Lemma}{Lemmas}

\newtheorem{theorem}{Theorem}

\newcommand{\ourmethod}{{\sc TextCause }}
\newcommand{\tboost}{{\textit{T-boost} }}
\newcommand{\wadj}{{\textit{W-Adjust} }}

\newcommand{\psireader}{\psi^\textrm{rea.}}
\newcommand{\psiwriter}{\psi^\textrm{wri.}}
\newcommand{\psinaive}{\psi^\textrm{naive}}
\newcommand{\psinaiveC}{\psi^\textrm{naive+C}}
\newcommand{\psiproxy}{\psi^\textrm{proxy}}

\newcommand{\psihatproxy}{\hat{\psi}^\textrm{proxy}}
\newcommand{\psihatnaive}{\hat{\psi}^\textrm{naive}}
\newcommand{\psihatc}{\hat{\psi}^\textrm{naive+C}}

\newcommand{\g}{\,\vert\,}
\newcommand{\s}{\,;\,}

\newcommand{\E}[1]{\mathbb{E}\left[#1\right]}

\newcommand{\rmdo}{\mathrm{do}}

\title{Causal Effects of Linguistic Properties}

\author{Reid Pryzant$^{1}$ \ \ Dallas Card$^1$ \ \ Dan Jurafsky$^1$ \ \ Victor Veitch$^2$ \ \ Dhanya Sridhar$^3$\\
  $^1$Stanford Unversity \\
  $^2$University of Chicago \\
  $^3$Columbia University\\  
  $^1$\texttt{\{rpryzant,dcard,jurafsky\}@stanford.edu}\\
    $^2$\texttt{victorveitch@gmail.com}\\
    $^3$\texttt{ds3778@columbia.edu}\\
}

\begin{document}
\maketitle


\begin{abstract}
We consider the problem of using observational data to estimate the causal effects of linguistic properties.
For example, does writing a complaint politely lead to a faster response time? How much will a positive product review increase sales? 
This paper addresses two technical challenges related to the problem before developing a practical method.
First, we formalize the causal quantity of interest as the effect of a \emph{writer's intent}, and establish the assumptions necessary to identify this from observational data.
Second, in practice, we only have access to noisy proxies for the linguistic properties of interest---e.g., predictions from classifiers and lexicons. We propose an estimator for this setting and prove that its bias is bounded when we perform an adjustment for the text. 
Based on these results, we introduce \textsc{TextCause}, an algorithm for estimating causal effects of linguistic properties. The method leverages (1) distant supervision to improve the quality of noisy proxies, and (2) a pre-trained language model (BERT)
to adjust for the text. We show that the proposed method outperforms related approaches when estimating the effect of Amazon review sentiment on semi-simulated sales figures. Finally, we present an applied case study investigating the effects of complaint politeness on bureaucratic response times.
\end{abstract}

\section{Introduction}

Social scientists have long been interested in the causal effects of language, studying questions like:
\begin{itemize}[noitemsep]
	\item How should political candidates describe their personal history to appeal to voters \citep{fong2016discovery2}?
	\item How can business owners write product descriptions to increase sales on e-commerce platforms \citep{pryzant2017predicting2,pryzant2018interpretable2}?
	\item How can consumers word their complaints to receive faster responses \citep{egami2018make}?
	\item What conversational strategies can mental health counselors use to have more successful counseling sessions \citep{zhang2020quantifying}?
\end{itemize}

To study the causal effects of linguistic properties, we must reason about interventions: what would the response time for a complaint be if we could make that complaint polite while keeping all other properties (topic, sentiment, etc.) fixed?
Although it is sometimes feasible to run such experiments where text is manipulated and outcomes are recorded \cite{newfonggrimmer}, analysts typically have observational data consisting of texts and outcomes obtained without intervention. 
This paper formalizes the estimation of causal effects of linguistic properties in observational settings.

Estimating causal effects from observational data requires addressing two challenges.
First, we need to formalize the causal effect of interest by specifying the hypothetical intervention to which it corresponds.
The first contribution of this paper is articulating the causal effects of linguistic properties; we imagine intervening on the writer
of a text document and telling them to use different linguistic properties.

The second challenge of causal inference is identification: we need to express causal quantities in terms of variables we can observe. Often, instead of the true linguistic property of interest we have access to a noisy measurement called the \emph{proxy label}. Analysts typically infer these values from text with classifiers, lexicons, or topic models \cite{grimmer2013text,lucas2015computer,prabhakaran2016predicting,voigt2017language,luo2019insanely,lucy2020content}.
The second contribution of this paper is establishing the assumptions we need to recover the true effects of a latent linguistic property from these noisy proxy labels. In particular, we propose an adjustment for the confounding information in a text document and prove
that this bounds the bias of the resulting estimates.

The third contribution of this paper is practical: an algorithm for estimating the causal effects of linguistic properties.
The algorithm uses distantly supervised label propagation to improve the proxy label \cite{zhur2002learning2,mintz2009distant,hamilton2016inducing}, then BERT to adjust for the bias due to text \cite{devlin2018bert,veitch2020adapting}.
We demonstrate the method's accuracy with partially-simulated Amazon reviews and sales data, perform a sensitivity analysis in situations where assumptions are violated, and show an application to consumer finance complaints. Data and a package for performing text-based causal inferences is available at \url{https://github.com/rpryzant/causal-text}.

\section{Causal Inference Background}
\label{sec:background}
Causal inference from observational data is well-studied \citep{pearl2009causality,rosenbaum1983central,rosenbaum1984reducing,shalizi2013advanced}. 
In this setting, analysts are interested in the effect of a \textbf{treatment} $T$ (e.g., a drug) on an \textbf{outcome} $Y$ (e.g., disease progression).
For ease, we consider binary treatments.
The average treatment effect (ATE) on the outcome $Y$ is,
\begin{align}
\psi = \E{Y \s \rmdo(T=1)} - \E{Y \s \rmdo(T=0)},
\end{align}
where the operation $\rmdo(T=t)$ means that we hypothetically intervene and set the treatment $T$ to some value \citep{pearl2009causality}.

Typically, the ATE $\psi$ is not the simple difference in average conditional outcomes, $\E{Y \g T=1} - \E{Y \g T=0}$. 
This is because \textbf{confounding} variables $C$ are associated with both the treatment and outcome, inducing
non-causal associations between them, referred to as \textbf{open backdoor paths} \cite{pearl2009causality}.
When all the confounding variables are observed, we can write the ATE in terms of observed variables using the 
\textbf{backdoor-adjustment formula} \cite{pearl2009causality},
 \begin{align}
 \label{eq:bd}
 \psi = \mathbb{E}_{C}\Big[\E{Y \g T=1, C} - \E{Y \g T=0, C}\Big].
 \end{align}
For example, if the confounding variable $C$ is discrete, 
we group the data into values of $C$, calculate the average difference in outcomes between the
treated and untreated samples of each group, and take the average over groups.

\section{Causal Effects of Linguistic Properties}
\label{sec:causal-model}

\begin{figure}[h!t]
	\centering
	\includegraphics[scale=0.15]{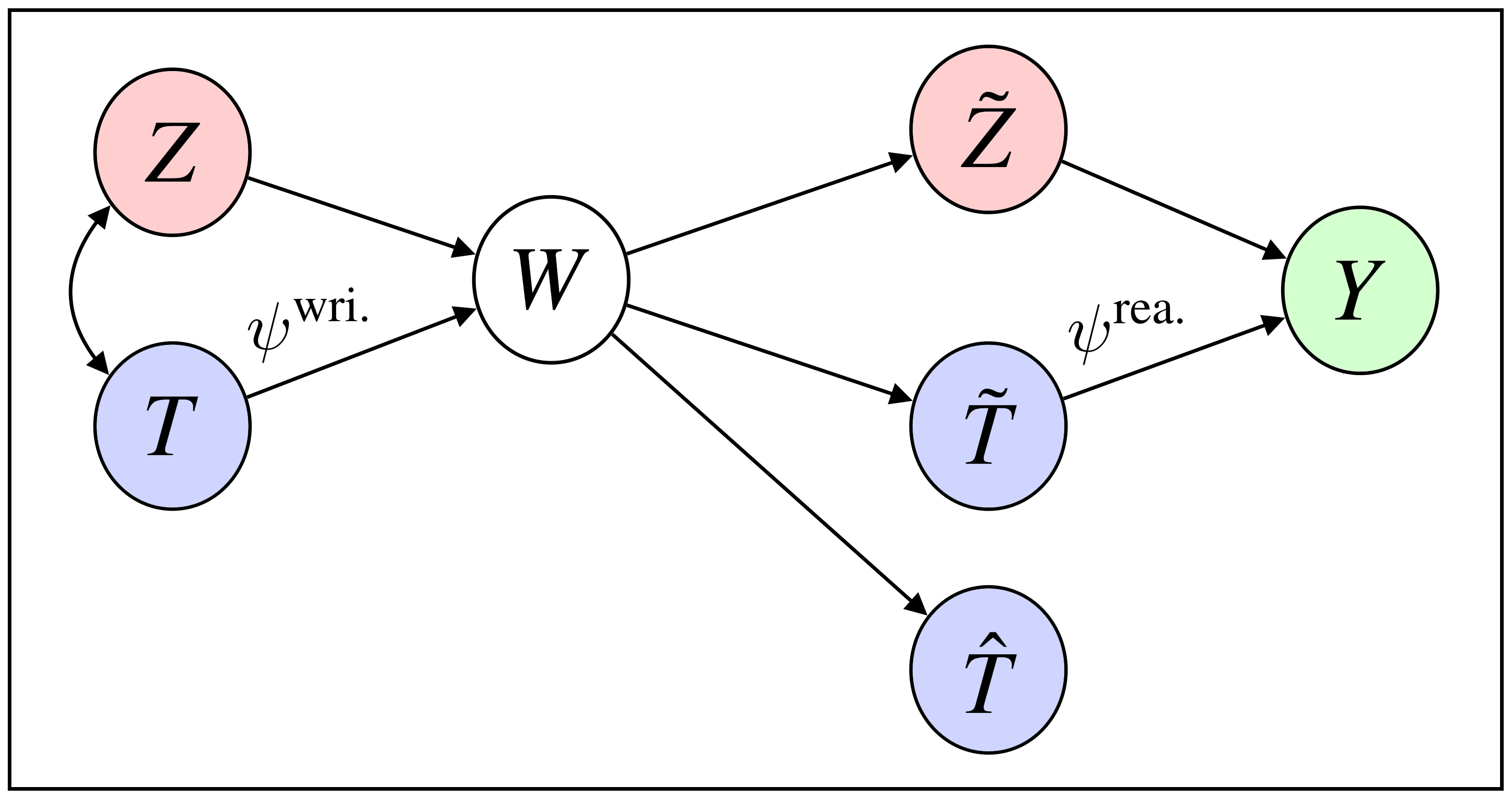}
	\caption{The proposed causal model of text and outcomes. A writer uses linguistic property $T$ and other properties $Z$, which may be correlated (denoted by bi-directed arrow), to write the text $W$. From the text, the reader perceives the property of interest, captured by $\tilde{T}$, and together with other perceived information $\tilde{Z}$, produces the outcome $Y$. The proxy label of the property obtained via a classifier or lexicon is captured by $\hat{T}$. 
		\label{fig:causal-gm}}
\end{figure}

We are interested in the causal effects of linguistic properties.  To formalize this as a treatment, we imagine intervening on the writer of a text, e.g., telling people to write with a property (or not). We show that to estimate the effect of using a linguistic property, we must consider how a reader of the text perceives the property. These dual perspectives of the reader and writer are well studied in linguistics and NLP;\footnote{Literary theory argues that language is subject to two perspectives: the ``artistic'' pole -- the text as intended by the author -- and the ``aesthetic'' pole -- the text as interpreted by the reader \cite{iser1974implied,iser1979act}. The noisy channel model \cite{yuret2010noisy,gibson2013rational}  connects these poles by supposing that the reader perceives a noisy version of the author's intent.
	This duality has also been modeled in linguistic pragmatics
	as the difference between speaker meaning and literal or utterance meaning
	\cite{potts2009formal,Levinson95,Levinson00}. Gricean pragmatic models like RSA \cite{goodman2016pragmatic} similarly formalize this as the reader using the literal meaning to help make inferences about the speaker's intent.} we adapt the idea for causal inference.

\Cref{fig:causal-gm} illustrates a causal model of the setting.
Let $W$ be a text document and let $T$ (binary) be whether or not a writer uses a particular linguistic property of interest.\footnote{We leave higher-dimensional extensions to future work.} For example,
in consumer complaints, the variable $T$ can indicate whether the writer intends to be polite or not.
The outcome is a variable $Y$, e.g., how long it took for this complaint to be serviced.
Let $Z$ be other linguistic properties that the writer communicated  (consciously or unconsciously) via the text $W$, e.g. topic, brevity or sentiment.
The linguistic properties $T$ and $Z$ are typically correlated, and both variables affect the outcome $Y$.

We are interested in the average treatment effect,
\begin{align}
\psiwriter = \E{Y \s \rmdo(T=1)} - \E{Y \s \rmdo(T=0)},
\end{align}
where we imagine intervening on writers and telling them to use the linguistic property of interest (setting $T=1$, ``\textit{write politely}'') or not ($T=0$).
This causal effect is appealing because the hypothetical intervention is well-defined -- it corresponds to an intervention we could perform in theory. However, without further assumptions, $\psiwriter$ is not identified from the observational data. The reason is that we would need to adjust for the unobserved linguistic properties $Z$, which create open backdoor paths because they are correlated with both the treatment $T$ and outcome $Y$ (\Cref{fig:causal-gm}).

To solve this problem, we observe that the \emph{reader} is the one who produces outcomes.
Readers use the text $W$ to perceive a value for the property of interest (captured by the variable $\tilde{T}$) as well as other properties (captured by $\tilde{Z}$) then produce the outcome $Y$ based on these perceived values. 
For example, a customer service representative reads a consumer complaint, judges whether (among other things) the complaint is polite or not, and chooses how quickly to respond based on this.

Consider the average treatment effect,
\begin{align}
\psireader = \E{Y \s \rmdo(\tilde{T}=1)} - \E{Y \s \rmdo(\tilde{T}=0)},
\end{align}
where we imagine intervening on the reader's perception of a linguistic property $\tilde{T}$.
The following result shows that we can identify the causal effect of interest, $\psiwriter$, by exploiting this ATE $\psireader$.
\begin{theorem}
\label{thm:identification}
   Let $\tilde{Z}=f(W)$ be a function of the words $W$ such that $\E{Y \g W} = \E{Y \g \tilde{T}, \tilde{Z}}$. 
	Suppose that the following assumptions hold:
	\begin{enumerate}
		\item (no unobserved confounding) $W$ blocks backdoor paths between $\tilde{T}$ and $Y$,
		\item (agreement of intent and perception) $T = \tilde{T}$.
		\item (overlap) For some constant $\epsilon > 0$, 
		$$\epsilon < P(\tilde{T}=1\g \tilde{Z}) < 1-\epsilon$$ with probability 1.\footnote{Informally, it must be possible to perceive a property ($\tilde{T}$=1) for all settings of $\tilde{Z}$, and $\tilde{Z}$ cannot perfectly predict $\tilde{T}$.}
	\end{enumerate}
Then the ATE $\psireader$ is identified as,
\begin{align}
\label{eq:text_adj}
\psireader = &\mathbb{E}_W\bigg[ \E{Y \g \tilde{T}=1, \tilde{Z}=f(W)} - \\
& \quad \quad \E{Y \g \tilde{T}=0, \tilde{Z}=f(W) }\bigg].
\end{align}
Moreover, the ATE $\psireader$ is equal to $\psiwriter$.
\end{theorem}
The proof is in Appendix A. 
Intuitively, the result says that the information in the text $W$ that the reader uses to determine the outcome $Y$ splits into two parts:
the information the reader uses to perceive the linguistic property of interest ($\tilde{T}$),
and the information used to perceive other properties ($\tilde{Z} = f(W)$).
The information captured by the variable $\tilde{Z}$ is confounding; it affects the outcome and is also correlated with the treatment $\tilde{T}$.
Under certain assumptions, adjusting for the function of text $\tilde{Z}$ that captures confounding suffices to identify the $\psireader$; in \Cref{fig:causal-gm}, the backdoor path $\tilde{T} \rightarrow W \rightarrow \tilde{Z} \rightarrow Y$ is blocked.\footnote{\citet{newfonggrimmer} studied a closely related setting
where text documents are randomly assigned to readers who produce outcomes.
From this experiment, they discover text properties that cause the outcome.
Their causal identification result requires an exclusion restriction assumption, which is related to the no unobserved confounding assumption that we make.}
Moreover, if we assume that readers correctly perceive the writer's intent,
the effect $\psireader$, which can be expressed in terms of observed variables, is equivalent to the effect
that we want, $\psiwriter$.


\section{Substituting Proxy Labels}
\label{sec:proxy-error}
If we observed $\tilde{T}$, the reader's perception of the linguistic property of interest, then we could proceed by estimating the effect $\psireader$ (equivalently, $\psiwriter$).
However, in most settings, one does not observe the linguistic properties that a writer intends to use ($T$ and $Z$) or that a reader perceives ($\tilde{T}$ and the information in $\tilde{Z}$).
Instead, one uses a classifier or lexicon to predict values for this property from the text, producing a proxy label $\hat{T}$ (e.g. predicted politeness). 

For this setting, where we only have access to proxy labels, we introduce the estimand $\psiproxy$ which substitutes the proxy $\hat{T}$ for the unobserved treatment $\tilde{T}$ in the effect $\psireader$:
\begin{align}
\label{eq:proxy}
\psiproxy =& \mathbb{E}_W\bigg[ \E{Y\g \hat{T}=1, \tilde{Z}=f(W)}\\
&\quad \quad - \E{Y\g \hat{T}=0, \tilde{Z}=f(W)} \bigg].
\end{align}
This estimand only requires an adjustment for the confounding information $\tilde{Z}$. We show how to extract this information using pretrained language models in Section \ref{section-methods}.
Prior work on causal inference with proxy treatments \citep{wood2018challenges} requires an adjustment using the measurement model $P(\tilde{T} \mid \hat{T})$, i.e. the true relationship between the proxy label $\hat{T}$ and its target $\tilde{T}$, which is typically unobserved.
In contrast, the estimand $\psiproxy$ does not require the measurement model.

The following result shows that the estimand $\psiproxy$ only attenuates the ATE that we want, $\psi^\textrm{rea.}$. That is, the bias due to proxy treatments is benign; it can only decrease the magnitude of the effect but it does not change the sign.

\begin{theorem}
	\label{thm:attenuation}
	Let $\epsilon_0 = \Pr(\tilde{T}=0\given \hat{T}=1,\tilde{Z})$ and let $\epsilon_1 = \Pr(\tilde{T}=1\given \hat{T}=0,\tilde{Z})$. Then,
	\begin{small}
		\begin{align*}
		\psiproxy = &\psi^\textrm{rea.} - 
		\mathbb{E}_W \Big[\big(\EE[Y\given \tilde{T}=1,\tilde{Z}] \\&-\EE[Y\given \tilde{T}=0,\tilde{Z}]\big)  \big(\epsilon_0 + \epsilon_1 \big)\Big]
		\end{align*}
	\end{small}
\end{theorem}
The proof is in Appendix E. 
This result shows that the proposed estimand $\psiproxy$, which we can estimate, is equal to the ATE $\psireader$ that we want, minus a bias term related to measurement error. 
In particular, if the classifier is better than chance and the treatment effect sign is homogeneous across possible texts --- i.e., it always helps or always hurts, an assumption the analyst must carefully assess --- then the bias term is positive with the degree of attenuation dependent on the error rate of the proxy label $\hat{T}$. The result tells us to construct the most accurate proxy treatment $\hat{T}$ possible, so long as we adjust for the confounding part of the text.\footnote{We prove in Appendix F 
that without the adjustment for confounding information $\tilde{Z}$, estimates of the ATE $\psireader$ will be arbitrarily biased.}
This is a novel result for causal inference with proxy treatments and sidesteps the need for the measurement model.


\begin{figure*}
    \centering
    \includegraphics[width=1\linewidth]{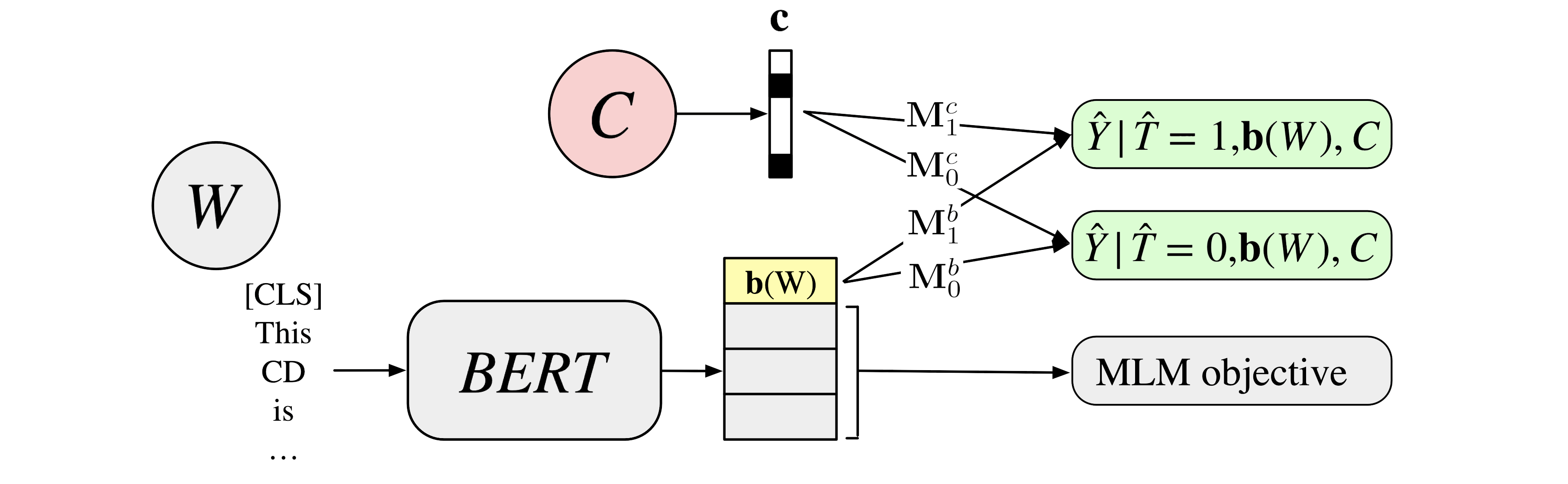}
    \caption{The second stage of \ourmethod adapts word embeddings to predict both of $Y$'s potential outcomes.}
    \label{fig:method1}
\end{figure*}

\section{\ourmethod, A Causal Estimation Procedure}
\label{section-methods}
We introduce a practical algorithm for estimating the causal effects of linguistic properties. Motivated by \Cref{thm:attenuation}, we first describe an approach for improving the accuracy of proxy labels. We then use the improved proxy labels, text and outcomes to fit a model that extracts and adjusts for the confounding information in the text $\tilde{Z}$. In practice, one may observe additional covariates $C$ that capture confounding properties, e.g., the product that a review is about or complaint type. We will include these covariates in the estimation algorithm.



\subsection{Improved Proxy Labels} 
\label{subsection-tstar}
The first stage of \ourmethod is motivated by \Cref{thm:attenuation}, which said that a more accurate proxy can yield lower estimation bias. Accordingly, this stage uses distant supervision to improve the fidelity of lexicon-based proxy labels $\hat{T}$. In particular, we exploit an inductive bias of frequently used lexicon-based proxy treatments: the words in a lexicon correctly capture the linguistic property of interest (i.e., high precision, \citealp{tausczik2010psychological}), but can omit words and discourse-level elements that also map to the desired property (i.e., low recall, \citealp{kim2006identifying,rao2009semi}).  

Motivated by work on lexicon induction and label propagation \cite{hamilton2016inducing,an2018semaxis}, we improve the recall of proxy labels, 
training a classifier $P_\theta$ to predict the proxy label $\hat{T}$, then using that classifier to relabel examples which were labeled $\hat{T}=0$ but look like $\tilde{T}=1$. Formally, given a dataset of tuples $\{(Y_i, W_i, C_i, \hat{T}_i)\}_{i=1}^n$ the algorithm is:
\begin{enumerate}
	\item Train a classifier to predict $P_\theta(\hat{T}\g W)$, e.g., logistic regression trained with bag-of-words features and $\hat{T}$ labels.
	\item Relabel some $\hat{T}=0$ examples (we experiment with ablating this in Appendix C): 
	\\$\hat{T}^{*}_i = \begin{cases}
	1 & \text{if }\hat{T}_i=1\\
	\mathbbm{1}[{P}_\theta(\hat{T}_i=1 \vert W_i) > 0.5] & \text{otherwise}
	\end{cases}$
	\item Use $\hat{T}^{*}$ as the new proxy treatment variable. 
\end{enumerate}

\subsection{Adjusting for Text}
\label{subsection-adjustW}

The second stage of \ourmethod estimates the effect $\psiproxy$ using the text $W$, improved proxy labels $\hat{T}^*$, and outcomes $Y$. This stage is motivated by \Cref{thm:identification}, which described how to adjust for the confounding parts of the text. We approximate this confounding information in the text, $\tilde{Z}=f(W)$, with a learned representation $\mathbf{b}(W)$ that predicts the expected outcomes $\mathbb{E}[Y\g \hat{T}^*=t, \mathbf{b}(W), C]$ for $t = 0, 1$ (\Cref{eq:proxy}).

We use DistilBERT \cite{sanh2019distilbert} to produce a representation of the text $\mathbf{b}(W)$ by embedding the text then selecting the vector corresponding to a prepended \texttt{[CLS]} token. We proceed to optimize the model so that the representation $\mathbf{b}(W)$ directly approximates the confounding information $\tilde{Z}=f(w)$. In particular, we train an estimator for the expected conditional outcome $Q(t, \mathbf{b}(W), C) = \mathbb{E}[Y \g \hat{T}^*=t, \mathbf{b}(W), C]$:
\begin{align}
\hat{Q}(t, \mathbf{b}(W), C) &= \sigma (\mathbf{M}^b_t \mathbf{b}(W) + \mathbf{M}^c_t \mathbf{c} + b )) \nonumber,
\end{align}
where the vector $\mathbf{c}$ is a one-hot encoding of the covariates $C$, the vectors $\mathbf{M}^b_t \in \mathbb{R}^{768}$ and $\mathbf{M}^c_t \in \mathbb{R}^{\vert C \vert}$ are learned, one for each value $t$ of the treatment, and the scalar $b$ is a bias term.

Letting $\theta$ be all parameters of the model, our training objective is to minimize,
\begin{align}
    \min_{\theta} \sum_{i=1}^n L(Y_i, \hat{Q}_\theta(\hat{T}^*_i,\mathbf{b}(W_i), C_i)) +  \alpha \cdot R(W_i) \nonumber,
\end{align}
where $L(\cdot)$ is the cross-entropy loss and $R(\cdot)$ is the original BERT masked language modeling objective, which we include following \citet{veitch2020adapting}. The hyperparameter $\alpha$ is a penalty for the masked language modeling objective. The parameters $\mathbf{M}_t$ are updated on examples where $\hat{T}^*_i=t$. 

Once $\hat{Q}(\cdot)$ is fitted, an estimator $\psihatproxy$ for the effect $\psiproxy$ (\Cref{eq:proxy}) is,
\begin{align}
\psihatproxy =&\frac{1}{n}\sum_i \Big[ \hat{Q}(1, \mathbf{b}(W_i), C_i) \nonumber \\
&\quad \quad \quad \quad- \hat{Q}(0, \mathbf{b}(W_i), C_i)  \Big],
\end{align}
where we approximate the outer expectation over the text $W$ with a sample average.
Intuitively, this procedure works because the representation $\mathbf{b}(W)$ extracts the confounding information $\tilde{Z} = f(W)$; it explains the outcome $Y$ as well as possible given the proxy label $\hat{T}^*$.

\section{Experiments}
\label{sec:experiments}

We evaluate the proposed algorithm's ability to recover causal effects of linguistic properties.
Since ground-truth causal effects are unavailable without randomized controlled trials, we produce a semi-synthetic dataset based on Amazon reviews where only the outcomes are simulated. We also conduct an applied study using real-world complaints and bureaucratic response times. Our key findings are
\begin{itemize}[noitemsep]
    \item More accurate proxies combined with text adjustment leads to more accurate ATE estimates.
    \item  Naive proxy-based procedures significantly underestimate true causal effects.
    \item ATE estimates can lose fidelity when the proxy is less than 80\% accurate. 
    \end{itemize}


\subsection{Amazon Reviews}
\label{subsection-amazon}

\subsubsection{Experimental Setup}

\textbf{Dataset.} Here we use real world and publicly available Amazon review data to answer the question, ``how much does a positive product review affect sales?'' We create a scenario where positive reviews increase sales, but this effect is confounded by the type of product. Specifically:

\begin{itemize}[noitemsep]
    \item The text $W$ is a publicly available corpus of Amazon reviews for digital music products \cite{ni2019justifying}. For simplicity, we only include reviews for mp3, CD, or Vinyl. We  also exclude reviews for products worth more than \$100 or fewer than 5 words.
    \item The observed covariate $C$ is a binary indicator for whether the associated review is a CD or not, and we use this to simulate a confounded outcome.
    \item The treatment $T = \tilde{T}$ is whether that review is positive (5 stars) or not (1 or 2 stars). Hence, we omit reviews with 3 or 4 stars. Note that here it is reasonable to assume writer's intention ($T$) equals the reader's perception ($\tilde{T}$), as the author is deliberately communicating their sentiment (or a very close proxy) with the stars. We use this variable to (1) simulate outcomes and (2) calculate ground truth causal effects for evaluation.
    \item The proxy treatment $\hat{T}$ is computed via two strategies: (1) a randomly noised version of $T$ fixed to 93\% accuracy (to resemble a reasonable classifier's output, later called ``\textbf{proxy-noised}''), and (2) a binary indicator for whether any words in $W$ overlap with a positive sentiment lexicon \cite{liu2010sentiment}. 
    \item The outcome $Y \sim \mathrm{Bernoulli} (\sigma( \beta_c (\pi(C) - \beta_o) + \beta_t \tilde{T} + \mathbb{N}(0, \gamma)))$  represents whether a product received a click or not. The parameter $\beta_c$ controls confound strength, $\beta_t$ controls treatment strength, $\beta_o$ is an offset and the propensity $\pi(C) = P(T = 1 \vert C)$ is estimated from data. 
\end{itemize}

The final data set consists of 17,000 examples. 

\textbf{Protocol.} All nonlinear models were implemented using PyTorch \cite{pytorch}. We use the \texttt{transformers}\footnote{\url{https://huggingface.co/transformers}} implementation of DistillBERT and the \texttt{distilbert-base-uncased} model, which has 66M parameters. To this we added 3,080 parameters for text adjustment (the $\mathbf{M^b_t}$ and $\mathbf{M^c_t}$ vectors). Models were trained in a cross-validated fashion, with the data being split into 12,000, 2,000, and 4,000-example train, validation, and test sets.\footnote{See \citet{egami2018make} for an investigation into train/test splits for text-based causal inference.} BERT was optimized for 3 epochs on each fold using Adam \cite{kingma2014adam}, a learning rate of $2e^{-5}$, and a batch size of 32. The weighting on the potential outcome and masked language modeling heads was 0.1 and 1.0, respectively. Linear models were implemented with \texttt{sklearn}. For T-boosting, we used a vocab size of 2,000 and L2 regularization with a strength of $c=1e^{-4}$. Each experiment was replicated using 100 different random seeds for robustness. Each trial took an average of 32 minutes with three 1.2 GHz CPU cores and one TITAN X GPU.

\textbf{Baselines.} The ``unadjusted'' baseline is $\psihatnaive = \hat{\EE}[Y \vert \hat{T} = 1] - \hat{\EE}[Y \vert \hat{T} = 0]$, the expected difference in outcomes conditioned on $\hat{T}$.\footnote{See Appendix F 
for an investigation into this estimator.} The  proxy-* baselines perform backdoor adjustment for the observed covariate $C$ and are based on \citet{sridhar2019estimating}:  $\psihatc  = \frac{1}{ \vert C \vert} \sum_{c}( \hat{\EE}[Y \vert \hat{T} = 1, C=c] - \hat{\EE}[Y \vert \hat{T} = 0, C=c]) $, using randomly drawn and lexicon-based $\hat{T}$ proxies. We also compare against ``semi-oracle'', $\hat{\psi}^{matrix}$, an estimator which assumes additional access to the ground truth measurement model  $P(\hat{T} \g T)$ \cite{wood2018challenges}; see Appendix G 
for derivation. 

\textbf{Note} that for clarity, we henceforth refer to the treatment-boosting and text-adjusting stages of \ourmethod as \tboost and \wadj.

\subsubsection{Results}

Our primary results are summarized in Table \ref{table-results1}. Individually, $\tboost$ and $\wadj$ perform well, generating estimates which are closer to the oracle than the naive ``unadjusted'' and ``proxy-lex' baselines. However, these components fail to outperform the highly accurate ``proxy-noised'' baseline unless they are combined (i.e., the \ourmethod algorithm). Only the full $\ourmethod$ algorithm consistently outperformed (i.e. produced  higher quality ATE estimates) than the baselines. This result is robust to varying levels of noise and treatment/confound strength. Indeed \ourmethod's estimates were on average within 2\% of the semi-oracle. Furthermore, these results  support \Cref{thm:attenuation}: methods which adjusted for the text always attenuated the true ATE. 

\begin{table*}[h!]
\begin{center}
\begin{tabular}{llrrrrrrrrc}
\hline \hline
\multicolumn{1}{r}{\textbf{Noise:}}       & \multicolumn{4}{c}{Low}                            & \multicolumn{4}{c}{High}                           \\
\multicolumn{1}{r}{\textbf{Treatment:}}   & \multicolumn{2}{c}{Low} & \multicolumn{2}{c}{High} & \multicolumn{2}{c}{Low} & \multicolumn{2}{c}{High} & Mean delta \\
\multicolumn{1}{r}{\textbf{Confounding:}} & Low        & High       & Low        & High        & Low        & High       & Low        & High & from oracle        \\ \hline \hline
oracle ($\psi$)       & 9.92 & 10.03 & 18.98 & 19.30 &   8.28 & 8.28 & 16.04 & 16.19 & 0.0 \\ 
semi-oracle ($\hat{\psi}^{matrix}$)  & 9.73 & 9.82 &  18.77  & 19.08 & 8.25 & 8.28 & 16.02 & 16.21 & 0.13 \\  \hline
unadjusted ($\psihatnaive$)   & 6.84 & 7.66 & 13.53 & 14.50 &     5.79 & 6.42 & 11.51 &  12.26 & 3.58 \\  
proxy-lex ($\psihatc $)  & 6.67 & 6.73 & 12.88  & 13.09 &    5.65 & 5.67 & 10.98 & 11.12 & 4.43 \\ 
proxy-noised ($\psihatc $) & 8.25 & 8.27 &  15.90  & 16.12 &   6.69 & 6.72 & 13.22 & 13.33 & 2.35 \\ \hline
+\tboost ($\psihatc $)    & 8.11 & 8.16 & 15.53 & 15.73 &     6.78 & 6.80 & 13.19 & 13.32 & 2.51 \\ 
+\wadj ($\psihatproxy$)       & 7.82 &  8.57  &  14.96  & 16.13 & 6.62 & 7.22 & 12.95 & 13.76 & 2.39\\
+\tboost +\wadj &   \textbf{9.42}  & \textbf{10.27} &  \textbf{18.20}  &  \textbf{19.32} & \textbf{7.85} & \textbf{8.53}& \textbf{15.45} & \textbf{16.30} & \textbf{0.37} \\ 
\ \ \ \  (\ourmethod, $\psihatproxy$) & & & & & & & & & \\ \hline
\hline
\end{tabular}
\vspace{-0.1cm}
\caption{ATE estimates: expected change in click probabilities if one were to manipulate the sentiment of a review from negative to positive.  $\textsc{TextCause}$ performs best in most settings. The true ATE is given in the top row (``oracle''). \textbf{Estimates closer to the oracle are better.} The last column gives the average difference between the estimated and true ATEs; lower is better. Rows 3-6 are baselines. Rows 7-9 are proposed. The second row and the bottom three rows use lexicon-based proxy treatments  (we observed similar results using other proxy treatments). All columns have $\beta_o=0.9$. Low and high noise corresponds to $\gamma$ = 0 and 1. Low and high treatment corresponds to $\beta_t$ = 0.4, 0.8. Low and high confounding corresponds to $\beta_c$ = -0.4, 4.0. All standard errors are less than 0.5. }
\label{table-results1}
\end{center}
\end{table*}

Our results suggest that adjusting for the confounding parts of text can be crucial: estimators that adjust for the covariates $C$ but not the text perform poorly, sometimes even worse than the unadjusted estimator $\psihatnaive$.

\textbf{Does it always help to adjust for the text?} 
We consider the case where confounding information in the text causes a naive estimator which does not adjust for this information ($\psinaive$) to have the opposite sign of the true effect $\psi$. Does our proposed text adjustment help in this situation? \Cref{thm:attenuation} says it should, because $\psiproxy$ estimates are bounded in [0, $\psi$]. This ensures that the most important of bits, the  bit of directional information, is preserved. 

\Cref{tab:crossing} shows results from such a scenario. We see that the true ATE of $T$, $\psi$, has a strong negative effect, while the naive estimator $\psinaiveC$ produces a positive effect. Adding an adjustment for the confounding parts of the text with \ourmethod successfully brings the proxy-based estimate to 0, which is indicative of the bounded behavior that \Cref{thm:attenuation} suggests.

\begin{figure}[h]
    \centering
    \includegraphics[width=1\linewidth]{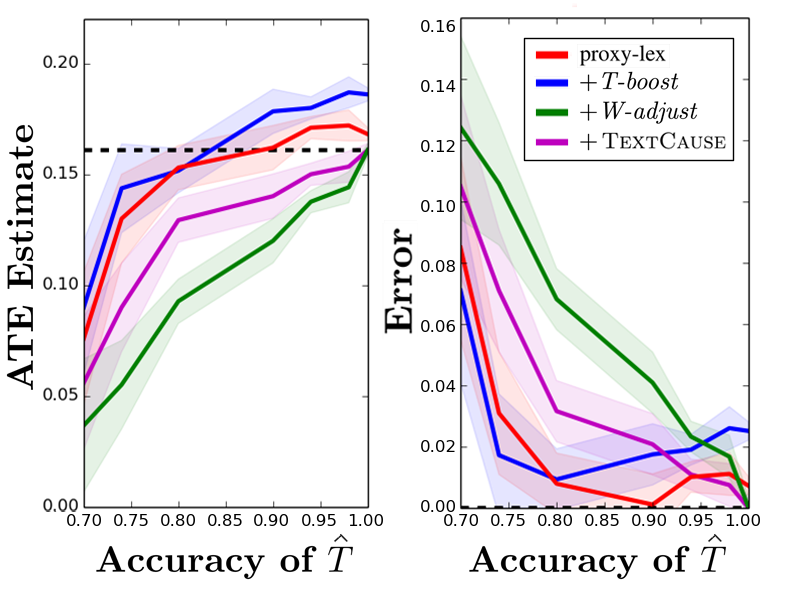}
    \vspace{-0.5cm}
    \caption{ ATE estimates as the accuracy of $\hat{T}$ is varied. Without text adjustment, \tboost's errors can increase with the error rate of $\hat{T}$.  The dotted black lines correspond to the true ATE (\textit{left}) and 0 error (\textit{right}).}
    \label{fig:crossing}
\end{figure}

\textbf{Sensitivity analysis.} In Figure \ref{fig:crossing} we synthetically vary the accuracy of a  proxy $\hat{T}$ by dropping random subsets of the data. This is to evaluate the robustness of various estimation procedures. We would expect (1) methods that do not adjust for the text to behave unpredictably, and (2) methods that do adjust for the text to be more robust.

\begin{table}[]
\centering
\begin{tabular}{l|ll}
\textbf{Estimator}                 & \textbf{ATE}    & \textbf{SE}  \\ \hline \hline
oracle ($\psi$)                   & -14.99 & $\pm$ 0.1 \\
proxy-lex  ($\psihatc $)            & 6.29   & $\pm$ 0.3 \\
+\tboost ($\psihatc $)    & 4.18   & $\pm$ 0.5 \\
+\tboost +\wadj & 0.50   & $\pm$ 1.3 \\
 \ \ (\ourmethod, $\psihatproxy$)
\end{tabular}
\vspace{-0.1cm}
\caption{ Estimator performance in a worst-case scenario where the estimated ATE of $\hat{T}$ and $\hat{T}^*$ indicates the opposite sign of the true ATE of $T$ ($\beta_c =0.8$, $\beta_t = -1, \pi(C) = 0.8$, $\beta_o = 0.6$).} 
\label{tab:crossing}
\end{table}

These results support our first hypothesis: boosting treatment labels without text adjustment can behave unpredictably, as proxy-lex and \tboost both overestimate the true ATE. In other words, the predictions of both estimators grow further from the oracle as $\hat{T}$'s accuracy increases. 

The results are mixed with respect to our second hypothesis. Both methods which adjust for the text (\wadj and \ourmethod) consistently attenuate the true ATE, which is in line with \Cref{thm:attenuation}. However, we find that \ourmethod, which makes use of $\tboost$ and \wadj, may not always provide the highest quality ATE estimates in finite data regimes. Notably, when $\hat{T}$ is less than 90\% accurate, both proxy-lex and \tboost can produce higher-quality estimates than the proposed \ourmethod algorithm.

Note that all estimates quickly lose fidelity as the proxy $\hat{T}$ becomes noisier. It rapidly becomes difficult for any method to recover the true ATE when the proxy $\hat{T}$ is less than 80\% accurate.

\subsection{Application: Complaints to the Financial Protection Bureau}
\label{subsection-cfpb}

We proceed to offer an applied pilot study which seeks to answer, ``how does the perceived politeness of a complaint affect the time it takes for that complaint to be addressed?'' We consider complaints filed with  the Consumer Financial Protection Bureau (CFPB).\footnote{\url{https://www.consumer-action.org/downloads/english/cfpb_full_dbase_report.pdf/}} This is a government agency which solicits and handles complaints about financial products. When they receive a complaint it is forwarded to the relevant company. The time it takes for that company to process the complaint is recorded. Some submissions are handled quickly ($<$ 15 days) while others languish. This 15-day threshold is our outcome $Y$. We additionally adjust for an observed covariate $C$ that captures what product and company the complaint is about (mortgage or bank account). To reduce other potentially confounding effects, we pair each $Y=1$ complaint with the most similar $Y=0$ complaint according to cosine similarity of TF-IDF vectors \cite{mozer2020matching}. From this we select the 4,000 most similar pairs for a total of 8,000 complaints. 

For our treatment (politeness), we use a state-of-the-art politeness detection package geared towards social scientists \cite{yeomans2018politeness}. This package reports a score from a trained classifier using expert features of politeness and a hand-labeled dataset. We take examples in the top and bottom 25\% of the scoring distribution to be our $\hat{T}=1$ and $\hat{T}=0$ examples and throw out all others. The final dataset consists of 4,000 complaints, topics, and outcomes.

We use the same training procedure and hyperparameters as Section \ref{subsection-amazon}, except now \wadj is trained for 9 epochs and each cross validation fold is of size 2,000.

\textbf{Results} are given in Figure \ref{tab:cfpb} and suggest that perceived politeness may have an effect on reducing response time. We find that the effect size increases as we adjust for increasing amounts of information. The ``unadjusted'' approach which does not perform any adjustment produces the smallest ATE. ``proxy-lex'', which only adjusts for covariates, indicated the second-smallest ATE. The $\wadj$ and \ourmethod methods, which adjust for covariates \emph{and} text, produced the largest ATE estimates. This suggests that there is a significant amount of confounding in real world studies, and the choice of estimator can yield highly varying conclusions. 

\begin{table}[]
\centering
\begin{tabular}{l|ll}
\textbf{Estimator} & \textbf{ATE} & \textbf{SE} \\ \hline  \hline
 unadjusted ($\psihatnaive$)  &  3.01    &  $\pm$ 0.3 \\
proxy-lex ($\psihatc $)     &  4.03    &  $\pm$ 0.4  \\
+\tboost ($\psihatc $)  &  9.64    &  $\pm$ 0.5  \\
+\wadj ($\psihatproxy$)      &  6.30    &  $\pm$ 1.6  \\
+\tboost +\wadj      &  10.30    & $\pm$ 2.1  \\
\ \ \ourmethod, ($\psihatproxy$ )
\end{tabular}
\vspace{-0.2cm}
\caption{Effect size can vary across estimation methods, with methods that adjust for more information producing larger ATEs. Each number represents the expected percent change in the likelihood of getting a timely response when the politeness of a complaint is hypothetically increased.}
\label{tab:cfpb}
\end{table}


\section{Related Work}

Our focus fits into a body of work on text-based causal inference that includes text as treatments \citep{egami2018make,fong2016discovery2,newfonggrimmer,wood2018challenges}, text as outcomes \cite{egami2018make}, and text as confounders (\citet{roberts.2020,veitch2020adapting}; see \citet{causalreview2020} for a review of that space). We build on \citet{veitch2020adapting}, which proposed a BERT-based text adjustment method similar to our \wadj algorithm.
This paper is related to work by \citet{newfonggrimmer}, which discusses assumptions needed to estimate causal effects of text-based treatments in randomized controlled trials.
There is also work on discovering causal structure in text, as topics with latent variable models \cite{fong2016discovery2} and as words and n-grams with adversarial learning \cite{pryzant2018deconfounded2} and residualization \cite{pryzant2018interpretable2}. There is also a growing body of applications in the social sciences \cite{hall2017,olteanu2017distilling,saha2019social,mozer2020matching,karell2019rhetorics,sobolev2018pro,zhang2020quantifying}.

This paper also fits into a long-standing body of work on measurement error and causal inference \citep{pearl2012measurement,kuroki2014measurement,buonaccorsi2010measurement,carroll2006measurement,shu2019weighted,oktay2019identifying,wood2018challenges}.
Most of this work deals with proxies for confounding variables. The present paper is most closely related to \citet{wood2018challenges}, which also deals with proxy treatments, but instead proposes an adjustment using the measurement model.

\section{Conclusion}
This paper addressed a setting of interest to NLP and social science researchers: estimating the causal effects of latent linguistic properties from observational data. We clarified critical ambiguities in the problem, showed how causal effects can be interpreted, presented a method, and demonstrated how it offers practical and theoretical advantages over the existing practice. We also release a package for performing text-based causal inferences.\footnote{\url{https://github.com/rpryzant/causal-text}}  This work opens new avenues for further conceptual, methodological, and theoretical refinement. This includes improving non-lexicon based treatments, heterogeneous effects, overlap violations, counterfactual inference, ethical considerations, extensions to higher-dimensional outcomes and covariates, and benchmark datasets based on paired randomized controlled trials and observational studies. 

\section{Acknowledgements}

This project recieved partial funding from the Stanford Data Science Institute, NSF Award IIS-1514268 and a Google Faculty Research Award. We thank Justin Grimmer, Stefan Wager, Percy Liang, Tatsunori Hashimoto, Zach Wood-Doughty, Katherine Keith, the Stanford NLP Group, and our anonymous reviewers for their thoughtful comments and suggestions.




\bibliographystyle{acl_natbib}
\bibliography{custom}

\onecolumn


\appendix

\section{Proof of Theorem 1}
\label{sec:thm1Proof}
Consider the expected outcome $\mu(t) = \E{Y \s \rmdo(\tilde{T}=t)}$. 
\begin{align}
    \mu(t) &= \mathbb{E}_W\bigg[\E{Y \g W \s \rmdo(\tilde{T}=t)}\bigg] \\ &\quad \textrm{(by iterated expectation)} \nonumber \\
    &= \mathbb{E}_W\bigg[\E{Y \g \tilde{T}, \tilde{Z} \s \rmdo(\tilde{T}=t)}\bigg] \\ &\quad \textrm{(by definition)} \nonumber\\
    &= \mathbb{E}_W\bigg[\E{Y \g \tilde{T}=t, \tilde{Z}}\bigg] \\ &\quad \textrm{(by overlap and no unobserved confounding)} \nonumber
\end{align}
The proof is complete because the estimand $\psireader$ is simply $\mu(1) - \mu(0)$.

\section{$C$-restriction}
\label{sec:crestrict}

Section \ref{sec:proxy-error} said that adjusting for confounding information $\tilde{Z} = f(W)$ is sufficient for blocking the confounding backdoor path created by non-treatment properties of the text that readers might perceive. In Section \ref{subsection-adjustW} we proposed performing this adjustment with BERT. We could alternatively try to block this path by restricting a $\hat{T}$-boosting model to only use information related to the confounding covariates $C$ (for which we can adjust). This could capture the same desired effect as conditioning on $\tilde{Z} = f(W)$ by accounting for whatever extra information was leaked from $W$ into $\hat{T}$. We accordingly experiment with restricting the model to features that are highly correlated with $C$: we compute point-biserial correlation coefficients \cite{glass1996statistical} between each word and $C$, then select the top 2000 words as features for the bootstrapping model. Results are given in Table \ref{table-climit} and suggest that while it still gives an improvement over the raw $\hat{T}$'s, $C$-restriction yields more conservative  estimates than adjusting for $W$. 

\begin{table}[H]
\begin{center}
\begin{tabular}{l|ccc}
                      & \textbf{Lexicon} & \textbf{BERT} & \textbf{Random} \\ \hline \hline
oracle ($\psi$)              & 15.54        &  15.54    &  15.54      \\
proxy-lex  ($\psihatc$)       & 11.21   &  7.83    & 12.34  \\
\tboost  &  13.30  &  10.92    &  12.70      \\
\ \ (\textit{C only}, $\psihatc$) & & & \\
\tboost ($\psihatc$)         &  14.68       &  12.01    &    13.59    \\
\wadj  ($\psiproxy$)           &  15.01       &  13.88    &  13.30     
\end{tabular}
\caption{ATE estimates from a $C$-restricted classifier (row 2) are preferable to $\hat{T}$ but conservative compared to less restrictive methods (\tboost and \wadj).} 
\label{table-climit}
\end{center}
\end{table}

\section{Ablating \tboost}
\label{sec:ablation}

\tboost only uses a classifier to change the treatment status of an example on $\hat{T}=0$ examples. Intuitively, this is because $\hat{T}$ is assumed to be a reasonable estimate of $\tilde{T}$ and therefore has a low false positive rate. We investigate this by ablating this part of the algorithm and directly setting $\hat{T}^{*}$ to the classifier's prediction. Our results on lexicon-based $\hat{T}$'s (Table \ref{table-ablation}, we observed similar outcomes with other $\hat{T}$'s) suggest this can reduce performance because a large number of correctly labeled $\hat{T}=1$ examples are flipped.

\begin{table}[H]
\centering
\begin{tabular}{l|c}
        \textbf{Estimator}                     & \textbf{Estimate} \\ \hline \hline
oracle  ($\psi$)                   & 15.54    \\
    proxy-lex  ($\psihatc$)                & 11.21    \\
\tboost ($\hat{T} = 0$ only, $\psihatc$) & \textbf{14.60}    \\
\tboost (all examples, $\psihatc$)   & 10.00    
\end{tabular}
\caption{It is advantageous to only relabel $\hat{T}=0$ examples.}
\label{table-ablation}
\end{table}

\section{Lemma 1 (used by Theorems 2 and 3)}
\label{lemma1}

\begin{small}
\begin{align*}
\EE[Y\given W, \hat{T}=1] & =\EE[Y\given W,T=1]\Pr(T=1\given W,\hat{T}=1) \\
&+ \EE[Y\given W, T=0]\Pr(T=0\given W,\hat{T}=1)\\
\EE[Y\given W, \hat{T}=0] & =\EE[Y\given W,T=0]\Pr(T=0\given W,\hat{T}=0) \\
&+\EE[Y\given W, T=1]\Pr(T=1\given W,\hat{T}=0)
\end{align*}
\end{small}
\begin{proof}
Apply the law of total probability and definition of conditional independence to the causal graph given in Figure 3. 
\end{proof}

\section{Proof of Theorem 2}
\label{sec:thm2}

Let
\begin{align*}
\epsilon_0 &= P(T=0 \g \hat{T} = 1,  \tilde{Z}) \\
\epsilon_1 &=  P(T=1 \g \hat{T} = 0,  \tilde{Z}, C) \\
p_1 &=  P(T=1 \g \hat{T} = 1,  \tilde{Z}) \\
p_0 &=  P(T=0 \g \hat{T} = 0,  \tilde{Z}) \\
E_1 &=  \mathbb{E}[Y \g \tilde{Z}, T=1] \\
E_0 &=  \mathbb{E}[Y \g \tilde{Z}, T=0] \\
\end{align*}
Now recall 
\begin{align*}
\psihatproxy &= \EE_{W}[\EE[Y \g \hat{T} = 1, \tilde{Z}] - \EE[Y \g \hat{T}=0, \tilde{Z}]] 
\end{align*}
Now we write the inner part using Lemma \ref{lemma1}, collect terms, and use the law of total probability to write everything in terms of misclassification probabilities: 
\begin{align*}
&= (E_1 p_1 + E_0 \epsilon_0) - (E_0 p_0 - E_1 \epsilon_1) \\
&= E_1 (p_1 + \epsilon_1) + E_0 (\epsilon_0 - p_0) \\
&= E_1 ((1 - \epsilon_0) +  \epsilon_1) + E_0 (\epsilon_0 - (1 - \epsilon_1)) \\
&= (E_1 - E_0)(1 - (\epsilon_0 + \epsilon_1))
\end{align*}
which completes the proof $\square$

\section{Theorem about the bias due to noisy proxies} 
\label{sec:thm3Proof}

Here we show that the naive estimand which does not adjust for the text,
\begin{align}
\psinaive = \E{Y \g \hat{T}=1} - \E{Y \g \hat{T}=0},
\end{align}
can be arbitrarily biased away from the effect of interest, $\psireader$.
\begin{theorem}
	\label{thm:bias}
		\begin{align*}
		\psinaive= \mathbb{E}_W \big[ \EE[Y\g \tilde{T}=1,W]\alpha(W) \ - \ \EE[Y\g \tilde{T}=0,W]\beta(W)\big]
		\end{align*}
	where
		\begin{align*}
		\alpha(W) & =\frac{P(\tilde{T}=1,\hat{T}=1\g W)}{P(\hat{T}=1)}-\frac{P(\tilde{T}=1,\hat{T}=0\g W)}{P(\hat{T}=0)}\\
		\beta(W) & =\frac{P(\tilde{T}=0,\hat{T}=0\g W)}{P(\hat{T}=0)}-\frac{P(\tilde{T}=0,\hat{T}=1\g W)}{P(\hat{T}=1)}
		\end{align*}
\end{theorem}

The $\alpha$ and $\beta$ terms are related to the error of the proxy label. This theorem says that correlations between the outcome and errors in the proxy can induce bias. Intuitively, this is similar to bias from confounding, though it is mathematically distinct. This means that even a highly accurate proxy label can result in highly misleading estimates. 
\emph{Proof:}
\begin{align*}
\EE[Y\given \hat{T}=1] & =\EE\big[\EE[Y\given \hat{T}=1,W]\given \hat{T}=1\big] \\
&=  \EE\left[\EE[Y\given \hat{T}=1,W]\frac{\Pr(W\given \hat{T}=1)}{\Pr(W)}\right]\\ 
&= \EE\left[\EE[Y\given \hat{T}=1,W]\frac{\Pr( \hat{T}=1\given W)}{\Pr( \hat{T}=1)}\right]
\end{align*}

Where the first equality is by the tower property, the second by inverse probability weighting, and the third Bayes' rule. We continue by invoking Lemma 1: 

\begin{align*}
\EE[Y\given \hat{T}=1,W]\frac{\Pr( \hat{T}=1\given W)}{\Pr( \hat{T}=1)} &= \EE[Y\given W,T=1]\Pr(T=1\given W, \hat{T}=1)\frac{\Pr( \hat{T}=1\given W)}{\Pr( \hat{T}=1)}\\
 &\quad\quad+\EE[Y\given W,T=0]\Pr(T=0\given W, \hat{T}=1)\frac{\Pr( \hat{T}=1\given W)}{\Pr( \hat{T}=1)} \\
 &= \EE[Y\given W,T=1]\frac{\Pr(T=1, \hat{T}=1\given W)}{\Pr( \hat{T}=1)}\\
  &\quad\quad+\EE[Y\given W,T=0]\frac{\Pr(T=0, \hat{T}=1\given W)}{\Pr( \hat{T}=1)}.
\end{align*}

The analogous expression for $\EE[Y\given \hat{T}=0]$:
\begin{align*}
\EE[Y\given \hat{T}=0] &=\EE \Bigg[\EE[Y\given W,T=0]\frac{\Pr(T=0, \hat{T}=0\given W)}{\Pr( \hat{T}=0)}\\
 & \quad\quad\quad+\EE[Y\given W,T=1]\frac{\Pr(T=1, \hat{T}=0\given W)}{\Pr( \hat{T}=0)}\Bigg]
\end{align*}
And now plugging into the ATE formula:
\begin{align*}
&\hat{\psi}^\textrm{rea.} = \EE[ \EE[Y ; do(\hat{T} = 1)] - \EE[Y ; do(\hat{T} = 0)] ] & \\
&\quad\ = \EE[ \EE[Y \vert \hat{T} = 1] - \EE[Y \vert \hat{T} = 0] ] \\
&\quad\ = \EE\Big[ \\
&\quad\quad\quad \EE[Y\given W,T=1]\Big( \frac{\Pr(T=1, \hat{T}=1\given W)}{\Pr( \hat{T}=1)}-\frac{\Pr(T=1, \hat{T}=0\given W)}{\Pr( \hat{T}=0)}\Big)\\
&\quad\quad\quad - \EE[Y\given W,T=0]\Big( \frac{\Pr(T=0, \hat{T}=0\given W)}{\Pr( \hat{T}=0)}-\frac{\Pr(T=0, \hat{T}=1\given W)}{\Pr( \hat{T}=1)}\Big) \\
& \quad\quad\quad\ \Big]
\end{align*}

The result follows immediately. $\square$

\section{Deriving semi-oracle, a Causal Estimator for when $P(T \vert \hat{T})$ is known}
\label{sec:matrix-adjust}

This is an ATE estimator which assumes access to $P(\hat{T}\vert T)$ instead of $T$ but is still unbiased. We derive this estimator using the ``matrix adjustment'' technique of   Wood-Doughty et al. (2018); Pearl(2012). We start by decomposing the joint distribution
\begin{align*}
    P(Y, T, \hat{T}, C) &= P(\hat{T} \vert Y, C, T) P(Y, C, T)\\
    P(Y, C, \hat{T}) &= \sum_T P(\hat{T} \vert Y, C, T) P(Y, C, T) 
\end{align*}

We can write this as a product between a matrix $\mathbf{M}_{c, y}(\hat{T}, T) = P(\hat{T} \vert Y, C, T)$ and vector $\mathbf{V}_{c, y}(T) =  P(Y, C, T)$:

\begin{align*}
\mathbf{V}_{c, y}(\hat{T}) &= \sum_T \mathbf{M}_{c, y}(\hat{T}, T) \mathbf{V}_{c, y}(T) \\
&= \mathbf{M}_{c, y} \mathbf{V}_{c, y}
\end{align*}

For our binary setting $\mathbf{M}_{c, y}$ is:
\begin{align*}
    \mathbf{M}_{c, y} &= \begin{bmatrix}
1- \delta_{c, y} & \epsilon_{c, y} \\
\delta_{c, y} & 1 - \epsilon_{c, y}
\end{bmatrix} \\
    \mathbf{M}_{c, y}^{-1} &= \frac{1}{1 - \epsilon_{c, y} - \delta_{c, y}} \begin{bmatrix}
1- \epsilon_{c, y} & -\epsilon_{c, y} \\
-\delta_{c, y} & 1 - \delta_{c, y}
\end{bmatrix}\\
\epsilon_{c, y} &= P(\hat{T}=0 \vert T=1, C, Y) \\
\delta_{c, y} &= P(\hat{T}= 1 \vert T=0, C, Y)
\end{align*}

Under fairly broad conditions, $\mathbf{M}$ has an inverse, which allows us to reconstruct the joint distribution:

\begin{align*}
    P(Y, T, C) = \sum_{\hat{T}} \mathbf{M}^{-1}_{c, y}(T, \hat{T}) \mathbf{V}_{c, y}(\hat{T})
\end{align*}

From which we can recover the ATE

\begin{align*}
    \psi^{matrix} &= \sum_c \left[ \frac{P(Y, T=1, C)}{\sum_Y P(Y, T=1, C)} - \frac{P(Y, T=0, C)}{\sum_Y P(Y, T=0, C)}  \right]\ P(C)
\end{align*}

Note also that this expression is similar to $\tau_{ME}$ in \citet{wood2018challenges} except their error terms are of the form $P(T \g \hat{T})$.


\end{document}